\documentclass[letterpaper]{article} % DO NOT CHANGE THIS
\usepackage{aaai2026}  % DO NOT CHANGE THIS
\usepackage{times}  % DO NOT CHANGE THIS
\usepackage{helvet}  % DO NOT CHANGE THIS
\usepackage{courier}  % DO NOT CHANGE THIS
\usepackage[hyphens]{url}  % DO NOT CHANGE THIS
\usepackage{graphicx} % DO NOT CHANGE THIS
\usepackage{natbib}  % DO NOT CHANGE THIS AND DO NOT ADD ANY OPTIONS TO IT
\usepackage{caption} % DO NOT CHANGE THIS AND DO NOT ADD ANY OPTIONS TO IT
\usepackage{algorithm}
\usepackage{algorithmic}
\usepackage[utf8]{inputenc} % allow utf-8 input
\usepackage[T1]{fontenc}    % use 8-bit T1 fonts
\usepackage{url}            % simple URL typesetting
\usepackage{booktabs}       % professional-quality tables
\usepackage{amsfonts}       % blackboard math symbols
\usepackage{nicefrac}       % compact symbols for 1/2, etc.
\usepackage{microtype}      % microtypography
 \usepackage{graphicx}
\usepackage{amsmath}
\usepackage{amssymb}
\usepackage{mathtools}
\usepackage{amsthm}
\usepackage{multirow}
\usepackage{multicol}
\usepackage{algorithm}
\usepackage{algorithmic}
\usepackage{graphicx}
\usepackage{array}
\usepackage{xcolor,colortbl}
\usepackage{xcolor}         % colors
\definecolor{tabhighlight}{HTML}{e5e5e5}
\usepackage[capitalize,noabbrev]{cleveref}
\usepackage{newfloat}
\usepackage{listings}
\DeclareCaptionStyle{ruled}{labelfont=normalfont,labelsep=colon,strut=off} % DO NOT CHANGE THIS
\floatstyle{ruled}
\newfloat{listing}{tb}{lst}{}
\floatname{listing}{Listing}
\title{Latent Knowledge-Guided Video Diffusion for Scientific Phenomena Generation from a Single Initial Frame}
\author{
    Qinglong Cao\textsuperscript{\rm 1,2}, Xirui Li\textsuperscript{\rm 1}, Ding Wang\textsuperscript{\rm 2}, Chao Ma\textsuperscript{\rm 1}, Yuntian Chen\thanks{Corresponding Author. Email: ychen@eitech.edu.cn},  Xiaokang Yang\textsuperscript{\rm 1}
}
\affiliations{
    \textsuperscript{\rm 1} MoE Key Lab of Artificial Intelligence, AI Institute, Shanghai Jiao Tong University, Shanghai, China\\
    \textsuperscript{\rm 2} Ningbo Institute of Digital Twin, Eastern Institute of Technology, Ningbo, China
}

\usepackage{bibentry}
\begin{document}

\maketitle

\begin{abstract}
Video diffusion models have achieved impressive results in natural scene generation, yet they struggle to generalize to scientific phenomena such as fluid simulations and meteorological processes, where underlying dynamics are governed by scientific laws. These tasks pose unique challenges, including severe domain gaps, limited training data, and the lack of descriptive language annotations. To handle this dilemma, we extracted the latent scientific phenomena knowledge and further proposed a fresh framework that teaches video diffusion models to generate scientific phenomena from a single initial frame. Particularly, static knowledge is extracted via pre-trained masked autoencoders, while dynamic knowledge is derived from pre-trained optical flow prediction. Subsequently, based on the aligned spatial relations between the CLIP vision and language encoders, the visual embeddings of scientific phenomena, guided by latent scientific phenomena knowledge, are projected to generate the pseudo-language prompt embeddings in both spatial and frequency domains.  By incorporating these prompts and fine-tuning the video diffusion model, we enable the generation of videos that better adhere to scientific laws. Extensive experiments on both computational fluid dynamics simulations and real-world typhoon observations demonstrate the effectiveness of our approach, achieving superior fidelity and consistency across diverse scientific scenarios.
\end{abstract}

% Uncomment the following to link to your code, datasets, an extended version or similar.
% You must keep this block between (not within) the abstract and the main body of the paper.
% \begin{links}
%     \link{Code}{https://aaai.org/example/code}
%     \link{Datasets}{https://aaai.org/example/datasets}
%     \link{Extended version}{https://aaai.org/example/extended-version}
% \end{links}

\section{Introduction}

Diffusion models, originally inspired by nonequilibrium thermodynamics~\cite{sohl2015deep}, have achieved remarkable success in various video generation tasks~\cite{saharia2022palette, kazerouni2022diffusion, he2025liangke, zhang2025decouple,he2024cameractrl}. The Video Diffusion Model (VDM)~\cite{ho2022video} pioneered the extension of diffusion models to video generation, yielding promising results. Subsequent works, such as Make-a-Video~\cite{singer2022make}, extended Text-to-Image (T2I) models to Text-to-Video (T2V) generation through super-resolution strategies. Further advancements, including Video LDM~\cite{blattmann2023align} and Imagen Video~\cite{ho2022imagen}, incorporated temporal consistency layers and cascaded generation architectures. Efficient training paradigms such as TAV~\cite{wu2023tune} have also been explored to reduce training overhead. Most recently, Stable Video Diffusion (SVD)~\cite{blattmann2023stable} and CogVideoX~\cite{yang2024cogvideox}, inspired by the principles of stable diffusion, has demonstrated superior performance through large-scale training.

Building upon these advancements, video diffusion models have begun to attract attention as general-purpose world simulators~\cite{yang2023learning}, which must internalize fundamental world laws to extrapolate beyond training data. However, recent studies~\cite{kang2024far} have shown that these models struggle to capture generalizable scientific principles. When applied to the scientific phenomena or systems governed by real-world scientific laws—such as dynamic fluids and typhoons—the performance of current video diffusion models degrades significantly. This is largely due to a pronounced domain gap and practical challenges like data scarcity and the absence of language annotations, which collectively hinder their ability to generate consistent dynamics.

 \begin{figure*}[h]
	\begin{center}
		%\fbox{\rule{0pt}{2in} \rule{0.9\linewidth}{0pt}}img
    \includegraphics[width=0.70\textwidth]{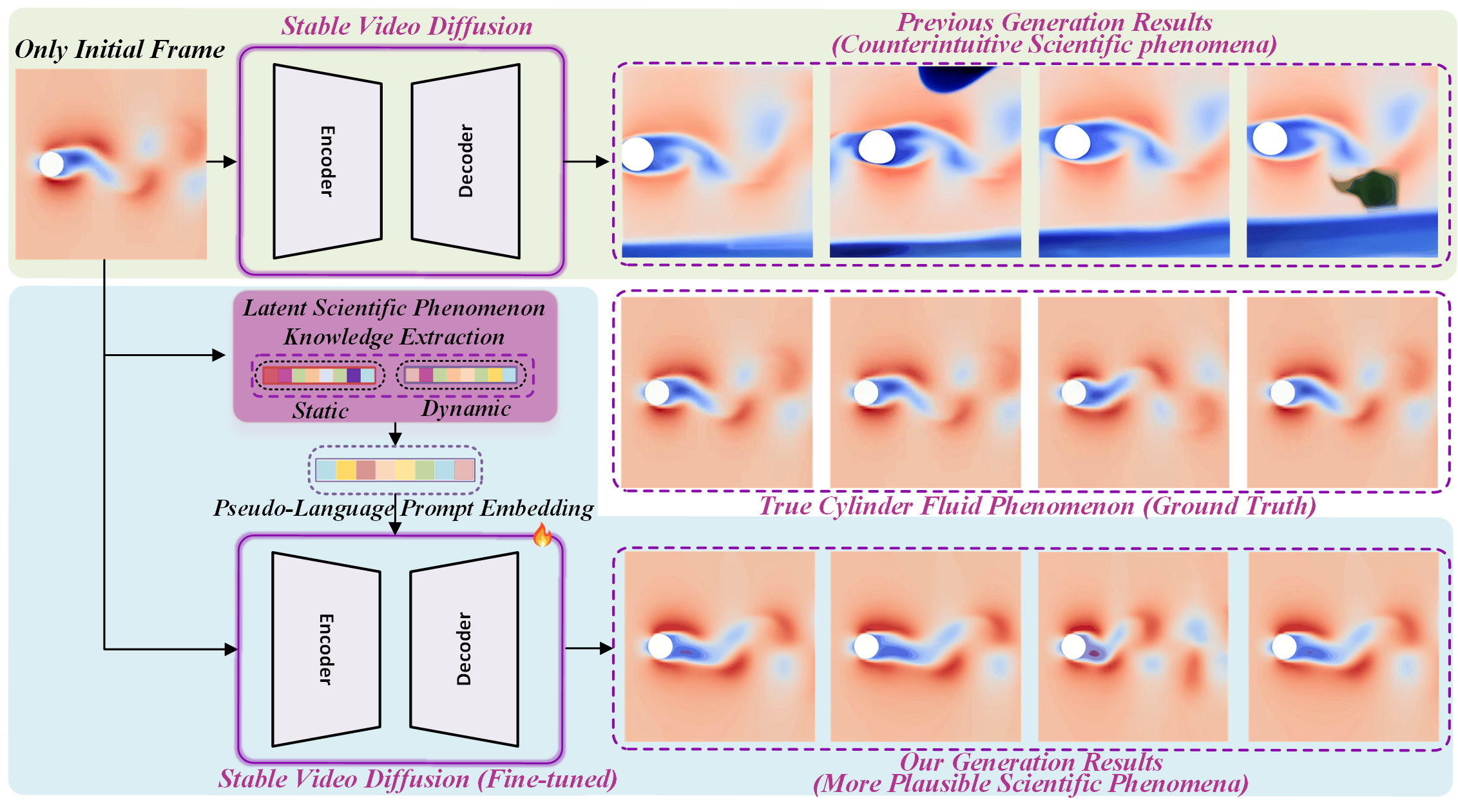}
	\end{center}
	\caption{Our approach integrates latent scientific phenomenon knowledge into video diffusion models via parameter-efficient fine-tuning, enabling more consistent and plausible generation under data-constrained scenarios.}
	\label{fig1}
\end{figure*}

To address these challenges, there is a growing need for video generation models that not only retain the generative power of diffusion architectures but also incorporate latent scientific phenomena knowledge to ensure alignment with real-world dynamics. To this end, we propose a novel framework that moves beyond traditional natural vision-based synthesis and toward latent knowledge-grounded scientific phenomena generation. Specifically, in scientific domains such as fluid mechanics and meteorology, realistic video generation requires models capable of generalizing from limited data availability, scientific consistency, and the absence of natural language prompts. Motivated by this need, our approach integrates latent scientific phenomenon knowledge into video diffusion models via parameter-efficient fine-tuning (Figure~\ref{fig1}), enabling more consistent and plausible generation under data-constrained scenarios.

In real data-scarce scientific environments, we assume access to only the first frame of a video. This raises a critical question: how can one extract meaningful latent scientific phenomenon knowledge from a single frame?~Self-supervised learning approaches~\cite{he2022masked,zhang2022dino,cao2025generalized} have demonstrated strong capabilities in capturing generalized visual representations, suggesting that their learned embeddings encode generalized latent knowledge. However, typical self-supervised augmentations such as noise injection, rotation, and color jitter are inappropriate for scientific data, where each pixel corresponds to a field variable with scientific meaning and directional force. These operations would violate the underlying scientific laws. To avoid such violations, we could only employ a Masked Autoencoder (MAE) to extract static latent scientific phenomenon knowledge. By reconstructing masked regions from unaltered observations, the MAE is implicitly encouraged to learn the governing scientific laws. While this provides a static understanding, scientific systems also involve dynamic behaviors. To capture dynamic knowledge, we leverage pre-trained optical flow predictors, which model the apparent motion patterns in scientific dynamic processes.

Although scientific knowledge can improve generative quality, existing video diffusion models often rely on language prompt embeddings to control generation. However, scientific phenomena are hard to articulate with descriptive natural language, making prompt formulation infeasible. Thus, we generate \emph{pseudo-language prompt embeddings} from the available visual data and scientific knowledge. This is enabled by the CLIP architecture, which offers well-aligned visual and language embedding spaces. We project the visual embeddings—enriched by latent scientific knowledge—into the language embedding space to produce pseudo-prompts. To capture complex multimodal dependencies, we adopt quaternion networks~\cite{shiprompt,cao2024domain} to perform the projection, which has shown effectiveness in cross-modal modeling of CLIP. Moreover, since frequency-domain information is key in representing scientific phenomena~\cite{zhang2024filtered, chen2025three}, we further inject spectral information to enrich the prompts.

Finally, we incorporate the pseudo-language prompt embeddings into video diffusion models through parameter-efficient fine-tuning, generating more scientifically realistic videos without relying on large-scale annotations or textual descriptions. Contributions are summarized as follows:

\begin{itemize}
\item We present the first framework for teaching video diffusion models to generate more scientifically plausible phenomena under practical constraints, including limited data and the absence of language annotations, by embedding latent scientific phenomenon knowledge.

\item We introduce a novel static-dynamic decomposition strategy that separates scientific phenomenon knowledge into static and dynamic components: static properties are preserved through masked autoencoding (MAE) to encourage scientifically consistent spatial representation, while dynamic properties are extracted via optical flow to capture temporal evolution patterns.
\item To overcome the challenge of lacking explicit language supervision, we design a quaternion-based projection module that transforms static and dynamic cues into pseudo-language prompt embeddings across spatial and frequency domains, enabling semantic-guided diffusion without natural language input.
\item Extensive experiments are performed on both numerical simulations and real-world observations of the scientific phenomena to validate the proposed method, demonstrating its promising performance across diverse scenarios.
\end{itemize}

\section{Related Work}
\textbf{Diffusion Models for Image Generation.} Diffusion models~\cite{ho2020denoising,song2020denoising}, rooted in nonequilibrium thermodynamic theory, have demonstrated remarkable success across a range of image synthesis tasks~\cite{croitoru2023diffusion, zhou2024image,gao2025auto}. Ho~\textit{et al.}~\cite{ho2022classifier} proposed classifier-free guidance to improve conditional generation quality, while Karras~\textit{et al.}~\cite{karras2022elucidating} investigated architectural refinements to advance generative fidelity. To accelerate inference, Salimans~\textit{et al.}~\cite{salimans2022progressive} introduced progressive distillation and alternative parameterizations. Beyond synthesis, diffusion-based methods have been applied to related domains including denoising~\cite{kawar2022denoising}, super-resolution~\cite{li2022srdiff}, and inpainting~\cite{lugmayr2022repaint}. Leveraging the capabilities of large-scale vision-language models~\cite{radford2021learning}, text-guided diffusion approaches~\cite{nichol2021glide, ramesh2022hierarchical} have improved image controllability. 

\textbf{Diffusion Models for Video Generation.} Compared to static images, video generation presents additional challenges due to temporal dynamics and motion continuity. Recent work has explored both adapting image diffusion models and training specialized architectures~\cite{xing2024simda,jiang2024videobooth}. Specifically, VideoLDM~\cite{blattmann2023align} extends latent diffusion to video through temporal tuning, while TAV~\cite{wu2023tune} adopts parameter-efficient strategies for text-to-video synthesis. Ho~\textit{et al.}~\cite{ho2022video} introduced a 3D extension of image diffusion for low-resolution video, and Stable Video Diffusion (SVD)~\cite{blattmann2023stable} later proposed a 3D U-Net to generate higher-resolution sequences. CogVideoX~\cite{yang2024cogvideox} utilizes expert transformer blocks to jointly process image and text tokens and reaches state-of-the-art video generation performance. Building upon SVD, Text2Video-Zero~\cite{khachatryan2023text2video} introduces cross-frame attention to enable zero-shot generation, while VidToMe~\cite{li2024vidtome}  refines inter-frame coherence via token-based fusion. Additionally, ControlVideo~\cite{zhang2023controlvideo}, inspired by ControlNet~\cite{zhang2023adding}, allows controllable video generation through structural priors.

\section{Preliminaries}
\textbf{Latent Diffusion Models.}
Diffusion models~\cite{sohl2015deep,ho2020denoising,song2020score} generate data by reversing a forward noising process: 
\begin{equation}
    x_t = \sqrt{\alpha_t} x_0 + \sqrt{1 - \alpha_t} \epsilon,
\end{equation}
where $\epsilon \sim \mathcal{N}(0, I)$ and $\{\alpha_t\}$ is a noise schedule. Latent diffusion models (LDMs)~\cite{rombach2022high,saharia2022photorealistic} operate in a compressed latent space via an autoencoder $z = \mathcal{E}(x)$, improving efficiency and quality. A UNet-based denoiser $\epsilon_\theta$ is trained to predict noise:
\begin{equation}
    \min_{\theta} \mathbb{E}_{z, \epsilon, t} \left[ \| \epsilon - \epsilon_\theta(z_t, t, c) \|^2 \right],
\end{equation}
where $c$ is optional conditioning (e.g., text). Our method builds upon video LDMs such as Stable Video Diffusion~\cite{blattmann2023stable} and CogVideoX~\cite{yang2024cogvideox}.

\textbf{Quaternion Neural Networks.}
A quaternion is $Q = r + x\mathbf{i} + y\mathbf{j} + z\mathbf{k}$ with $r,x,y,z \in \mathbb{R}$. Quaternion multiplication (Hamilton product) preserves 3D rotational and spatial structure. In quaternion neural networks (QNNs)~\cite{parcollet2018quaternion}, layers compute:
\begin{equation}
Q_{\text{out}} = \alpha(W \otimes Q),
\end{equation}
where $W$ is a quaternion weight and $\alpha$ applies an activation (e.g., ReLU) component-wise:
\begin{equation}
\alpha(Q) = f(r) + f(x)\mathbf{i} + f(y)\mathbf{j} + f(z)\mathbf{k}.
\end{equation}
QNNs inherently model multi-dimensional correlations, making them suitable for spatially aligned multimodal fusion—unlike real-valued networks that often lose geometric structure.

\section{Method}

 \begin{figure*}[t!]
	\begin{center}
		%\fbox{\rule{0pt}{2in} \rule{0.9\linewidth}{0pt}}img
		\includegraphics[width=0.72\linewidth]{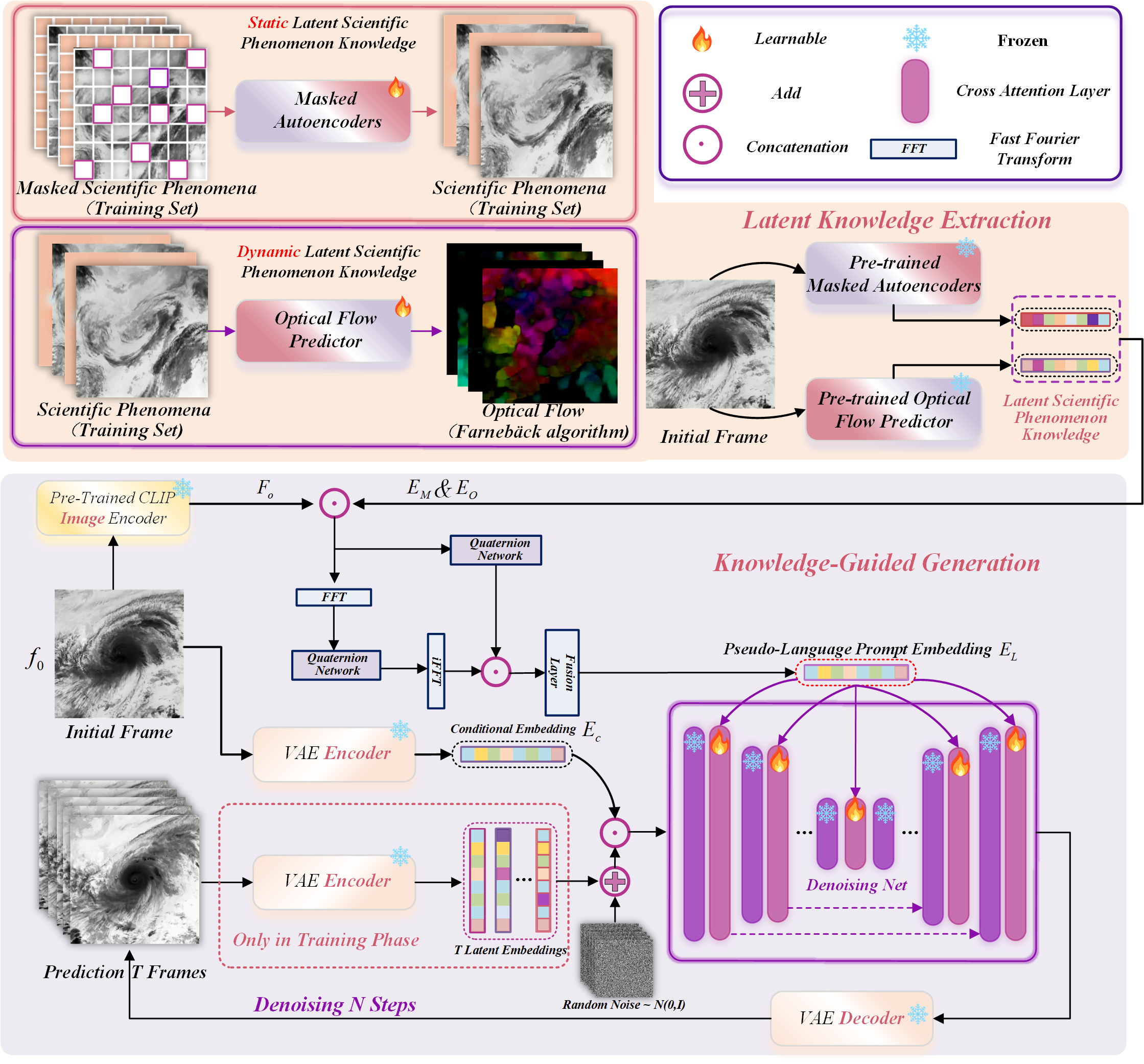}
	\end{center}
	\caption{Overview of our proposed method. Using the MAE and optical flow prediction to extract latent physical phenomenon knowledge. Projecting CLIP vision features guided by latent physical phenomenon knowledge to obtain pseudo-language prompt embeddings. Incorporating these embeddings to generate more physically plausible physical phenomena. 
 }
	\label{fig2}
\end{figure*}

\subsection{Video Generation Pipeline}
As illustrated in Figure~\ref{fig2}, our framework generates scientific phenomenon videos from a single initial frame via two stages: (1) latent knowledge extraction and (2) knowledge-guided generation. From raw training videos (without annotations), we extract static knowledge using a Masked Autoencoder (MAE) and dynamic motion via an Optical Flow Predictor (OFP) trained on pseudo ground-truth flows~\cite{farneback2003two}. Both modules are frozen after training.
Given an initial frame \( f_0 \), MAE and OFP produce static and dynamic embeddings \( E_M \) and \( E_O \). Meanwhile, a pre-trained VAE and CLIP encoder yield conditional features \( E_c \) and \( F_o \). \( F_o \) is concatenated with \( E_M \) and \( E_O \), then projected into quaternion space to form pseudo-language prompt embeddings \( E_L \), which are injected into the UNet’s cross-attention layers. The UNet is fine-tuned using LoRA~\cite{hu2021lora}.
During training, all frames of the target video \( \mathcal{V}_p \) are encoded into \( T \) latents, noised to form \( N_{\text{input}} \), and the UNet learns denoising via iterative reconstruction. At inference, frozen MAE and OFP re-extract \( E_M \) and \( E_O \) from \( f_0 \); \( E_c \) is combined with sampled noise \( N_{\text{input}} \) to form conditioned input \( N'_{\text{input}} \), enabling video synthesis guided by both visual cues and \( E_L \).

\subsection{Latent Knowledge Extraction}
% \subsection{Latent Physical Phenomenon Knowledge}
Given $K$ training scientific phenomena $[\mathcal{V}_1, \mathcal{V}_2, \dots, \mathcal{V}_K]$, we decompose each video into individual frames to construct an image set $[I_1^{V_1}, I_2^{V_1}, \dots, I_N^{V_K}]$.
To extract static latent scientific phenomenon knowledge, each image is partitioned into non-overlapping, regular patches to facilitate processing. Then, since self-supervised learning methods like SimCLR or DINO rely on augmentations (e.g., rotation, noise addition) that would violate the underlying scientific law, we adopt a Masked Autoencoder (MAE), which operates on non-overlapping image patches. A subset of patches $P_{\text{vis}}$ is randomly retained as visible tokens, while the remaining $P_{\text{masked}}$ are withheld to promote spatial prediction. A Vision Transformer (ViT)~\cite{dosovitskiy2020image} serves as both encoder and decoder of the MAE, trained to reconstruct the masked regions from visible context:
\begin{equation}
    P_{\text{masked}} = \text{MAE}(P_{\text{vis}}).
\end{equation}
The reconstruction is optimized via mean squared error (MSE). By reconstructing masked regions from unaltered observations, the MAE is implicitly encouraged to learn the governing scientific laws.
For dynamic latent knowledge, we utilize optical flow to capture inter-frame motion. Pseudo ground-truth flows are computed via the Farnebäck algorithm~\cite{farneback2003two}, and used to supervise an Optical Flow Predictor (OFP), which also adopts a ViT backbone. Given an input frame $I_{\text{input}}$, the OFP is trained to estimate the flow $f_{\text{op}}$:
\begin{equation}
    f_{\text{op}} = \text{OFP}(I_{\text{input}}).
\end{equation}
The MSE loss again guides the training, enabling the OFP to encode dynamic evolution patterns.
Upon convergence, we extract static and dynamic knowledge embeddings $E_M$ and $E_O$ of the only acquirable initial frame from the decoder outputs of the frozen MAE and OFP, respectively. 
% These latent representations capture complementary aspects—static and dynamic knowledge—of scientific phenomena, forming rich knowledge priors for video generation.

\subsection{Knowledge-Guided Generation}
We first encode the initial frame $f_0$ using a pre-trained VAE and CLIP image encoder to obtain conditional features $E_c$ and visual features $F_o$. Unlike traditional video diffusion models that predominantly rely on natural language prompts, our method targets more realistic scenarios where describing scientific phenomena in natural language is often intractable. While prior models commonly use CLIP’s text encoder to generate language embeddings, we instead leverage CLIP’s robust vision-language alignment to derive pseudo-language prompt embeddings directly from visual features and extracted latent knowledge embeddings.
To this end, we incorporate static and dynamic latent physical knowledge, denoted as $E_M$ and $E_O$, respectively. Inspired by the success of prior works~\cite{shiprompt,cao2024domain} that utilize quaternion networks~\cite{parcollet2018quaternion} for modeling multimodal CLIP's relations, we adopt quaternion networks to derive pseudo-language prompt embeddings. We first apply linear layers $[L_{d1}, L_{d2}, L_{d3}]$ to obtain projected features:
\begin{equation}
\hat{F}_o = L_{d1}(F_o),\hat{E}_M = L_{d2}(E_M),\hat{E}_O = L_{d3}(E_O).
\end{equation}
We randomly initialize learnable text embeddings $T_L$ and organize the four components—text, vision, static knowledge, and dynamic knowledge—along the orthogonal axes of the quaternion latent space:
\begin{equation}
Q_l = T_L + \hat{F}_o\mathbf{i} + \hat{E}_M\mathbf{j} + \hat{E}_O\mathbf{k}.
\end{equation}
As frequency-domain information are critical for modeling scientific dynamics~\cite{zhang2024filtered,chen2025three}, we extend this formulation using the Fast Fourier Transform (FFT):
\begin{equation}
Q_l^{FFT}=T_L^{FFT} + FFT(\hat{F}_o)\mathbf{i} + FFT(\hat{E}_M)\mathbf{j} + FFT(\hat{E}_O)\mathbf{k},
\end{equation}
where $T_L^{FFT}$ denotes learnable embeddings in the frequency domain. These quaternion latent vectors are processed by different quaternion networks $Q_t$ and $Q_t^{FFT}$ to generate pseudo-language prompt embeddings:
\begin{equation}
\begin{split}
E_L^S &= Q_t([T_L, \hat{F}_o, \hat{E}_M, \hat{E}_O]),\\
E_L^F &= Q_t^{FFT}([T_L^{FFT}, FFT(\hat{F}_o), FFT(\hat{E}_M), FFT(\hat{E}_O)]).
\end{split}
\end{equation}
A fusion network $L_{\text{fuse}}$, consisting of two linear layers, combines spatial and frequency-domain features to form the final prompt embeddings:
\begin{equation}
E_L = L_{\text{fuse}}([E_L^S, E_L^F]).
\end{equation}
These embeddings $E_L$ are then injected into each cross-attention layer of the stable video diffusion UNet via parameter-efficient fine-tuning. Let $\varphi_i(z_t) \in \mathbb{R}^{N \times d_\epsilon^i}$ be the projected UNet hidden state at layer $i$ and $\tau_\theta(E_L) \in \mathbb{R}^{M \times d_\tau}$ be the prompt representation. The attention mechanism is:
\begin{equation}
\text{Attention}(Q, K, V) = \text{softmax}\left(\frac{QK^T}{\sqrt{d}}\right)V.
\end{equation}
We apply LoRA~\cite{hu2021lora} to the projection matrices:
\begin{equation}
\begin{split}
Q &= \text{LoRA}(W_Q^{(i)}) \cdot \varphi_i(z_t),\\
K &= \text{LoRA}(W_K^{(i)}) \cdot \tau_\theta(E_L),\\
V &= \text{LoRA}(W_V^{(i)}) \cdot \tau_\theta(E_L),
\end{split}
\end{equation}
where $W_Q^{(i)}, W_K^{(i)}, W_V^{(i)}$ are frozen pretrained weights with trainable LoRA adapters.
In summary,  the  video diffusion model \(D_t\), combined with the CLIP image encoder \(E_{\textit{img}}\) and the quaternion projection process \(Q\) applied in both spatial and frequency domains, enables robust video generation:
\begin{equation}
\begin{split}
E_L &= Q(E_{\textit{img}}(f_0), \text{MAE}(f_0), \text{OFP}(f_0)),\\
\mathcal{V}_p &= D_t([\text{DDIM}_{\text{sample}}, f_0], E_L),
\end{split}
\end{equation}
where $f_0$ is the initial frame and $\mathcal{V}_p$ is the generated video.

 \begin{figure*}[t!]
	\begin{center}
		%\fbox{\rule{0pt}{2in} \rule{0.9\linewidth}{0pt}}
		\includegraphics[width=0.70\linewidth]{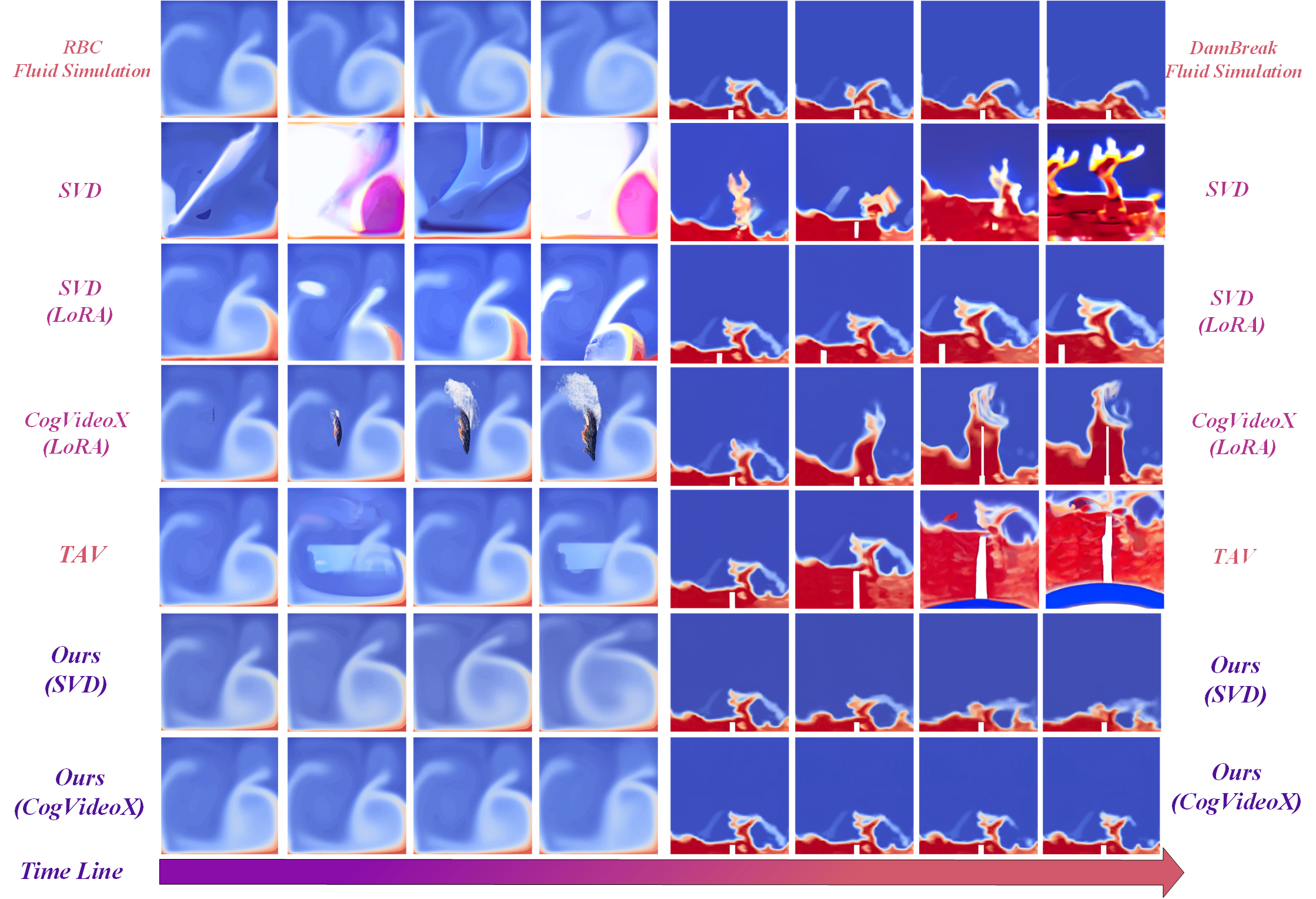}
	\end{center}
	\caption{Qualitative results in fluid simulation datasets. Our method, guided by latent physical knowledge, produces phenomena more consistent with physical laws.}
	\label{fig3}
\end{figure*}

 \begin{figure*}[t!]
	\begin{center}
		%\fbox{\rule{0pt}{2in} \rule{0.9\linewidth}{0pt}}
		\includegraphics[width=0.70\linewidth]{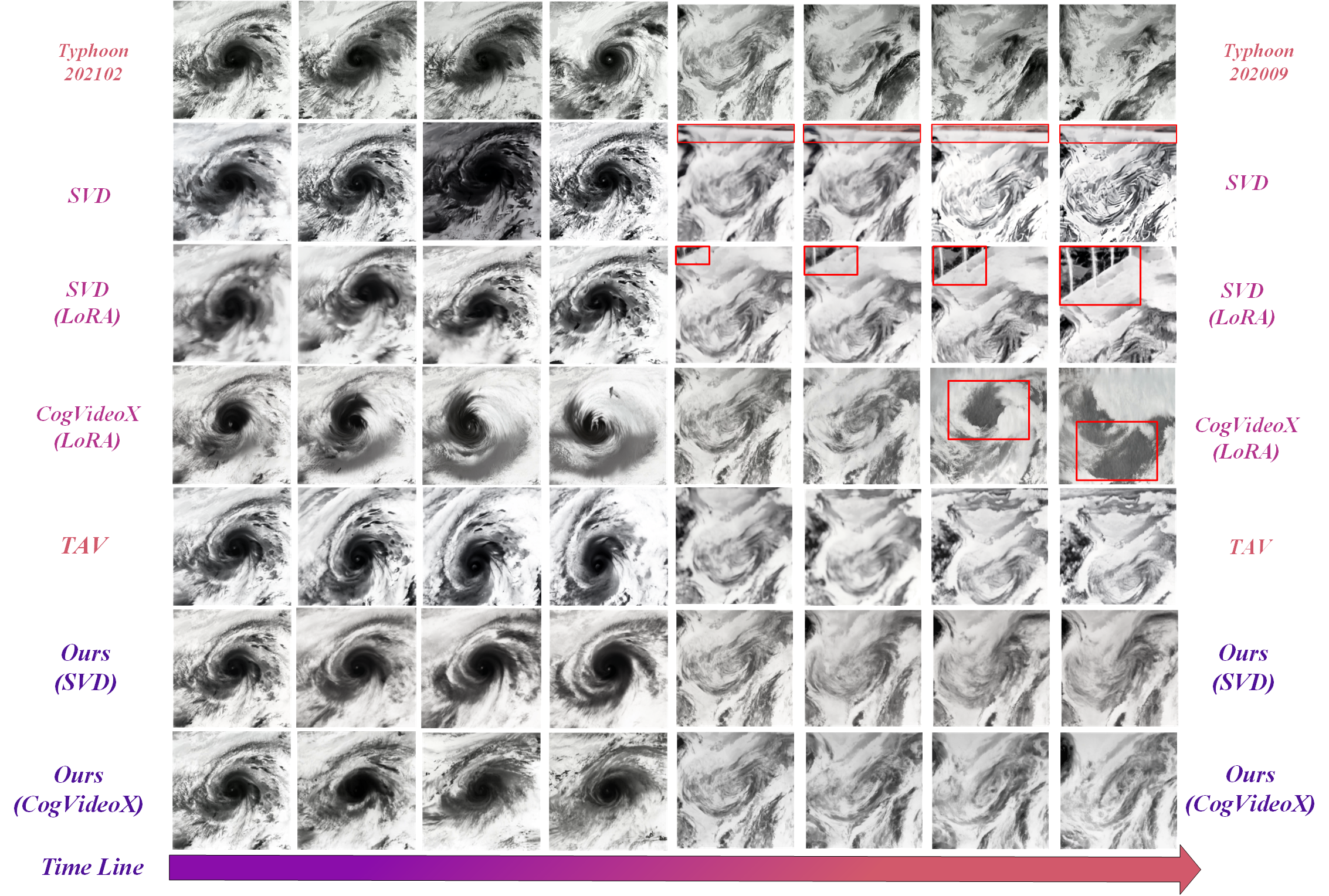}
	\end{center}
	\caption{Qualitative comparisons in true typhoon dataset. The red box denotes some hallucinations.}
	\label{fig4}
\end{figure*}

\section{Experiments}
\subsection{Experiment Setting}
We adopt Stable Video Diffusion (SVD)~\cite{blattmann2023stable} and CogVideoX~\cite{yang2024cogvideox} as the baseline and evaluate methods on both simulated fluid dynamics and real-world typhoon datasets. We employ a mix of numerical accuracy metrics and physically grounded evaluation criteria to ensure a comprehensive assessment.

\begin{table*}[t!]
    \scriptsize
    \footnotesize
    \centering
    \renewcommand{\arraystretch}{0.8}
    \renewcommand{\tabcolsep}{1.6mm}
    \scalebox{0.75}{
    % \begin{tabular}{lcccccccc|cccccc}
    \begin{tabular}{@{}p{1.8cm}@{}cccccccc|cccccc}
    \toprule
    & RMSE$\downarrow$ & SSIM$\uparrow$ & SFE$\downarrow$ & SE$\downarrow$ & GS$\downarrow$ & CS$\downarrow$ & QCE$\downarrow$ & VE$\downarrow$ & RMSE$\downarrow$ & SSIM$\uparrow$ & SFE$\downarrow$ & SE$\downarrow$ & GS$\downarrow$ & CS$\downarrow$ \\ 
    \midrule
    \multicolumn{9}{c|}{\textbf{\textit{RBC fluid simulation}}} &
    \multicolumn{6}{c}{\textbf{\textit{Typhoon 202001}}}  \\ \midrule
    SVD & 9.323 & 0.773 & 170.48 & 1.893 & 4.826 & 12.46 & 0.075 & 0.029 & 9.702 & 0.365 & 1575.8 & 2.942 & 20.26 & \textcolor{red}{\textbf{16.66}} \\ 
    % CogX &8.618 &  0.899 & 4.4674 & 0.0999 & 1.8795 & 0.8598 & 0.0011 & 0.0002 &  &  &  &  &  &  \\
    LoRA(SVD) & 8.876 & 0.831 & 82.159 & 0.609 & 3.670 & 5.773 & 0.018 & 0.004 & 9.642 & 0.395 & 2733.6 & 7.240 & 17.32 & 31.21 \\
    LoRA(CogX) & 8.885 &  0.871 & 222.46 &  0.971 & 5.2284 & 10.29 & 0.068 & 0.010 &  9.909 & 0.411 & 2966.7 & 14.39 & 10.73 & 51.50 \\
    TAV & 8.457 & 0.863 & 15.470 & 0.229 & 2.731 & 1.706 & 0.003 & 6e-4 & 9.696 & 0.367 & 3297.7 & 15.53 & 18.94 & 37.20 \\
    SimDA & 8.786 & 0.857 & 14.895 & 0.256 & 3.012 & 1.716 & 0.003 & 8e-4 & 9.710 & 0.375 & 2846.3 & 14.67 & 18.37 & 34.61 \\
        \rowcolor{tabhighlight}
    Ours(SVD) & \textcolor{blue}{\textbf{8.490}} & \textcolor{blue}{\textbf{0.902}} & \textcolor{blue}{\textbf{4.4569}} & \textcolor{blue}{\textbf{0.099}} & \textcolor{blue}{\textbf{2.081}} & \textcolor{blue}{\textbf{0.819}} & \textcolor{blue}{\textbf{0.001}} & \textcolor{blue}{\textbf{2e-4}} & \textcolor{blue}{\textbf{9.351}} & \textcolor{blue}{\textbf{0.446}} & \textcolor{blue}{\textbf{606.36}} & \textcolor{red}{\textbf{1.346}} & \textcolor{blue}{\textbf{10.90}} & \textcolor{blue}{\textbf{19.19}} \\
     \rowcolor{tabhighlight}
   Ours(CogX) & \textcolor{red}{\textbf{7.933}} & \textcolor{red}{\textbf{0.911}} & \textcolor{red}{\textbf{4.2683}} & \textcolor{red}{\textbf{0.064}} & \textcolor{red}{\textbf{1.893}} & \textcolor{red}{\textbf{0.459}} & \textcolor{red}{\textbf{7e-4}} & \textcolor{red}{\textbf{2e-4}} & \textcolor{red}{\textbf{9.314}} & \textcolor{red}{\textbf{0.502}} & \textcolor{red}{\textbf{1063.5}}  & \textcolor{blue}{\textbf{2.531}} & \textcolor{red}{\textbf{9.167}} &  28.27\\

    \midrule
    \multicolumn{9}{c|}{\textbf{\textit{Cylinder fluid simulation}}} &
    \multicolumn{6}{c}{\textbf{\textit{Typhoon 202009}}} 
  \\ \midrule
    SVD & 8.081 & 0.871 & 399.88 & 1.453 & 4.832 & 15.83 & 0.211 & 0.022 & 10.29 & 0.308 & 3443.9 & 7.828 & 18.94 & 32.44 \\ 
    LoRA(SVD) & 7.206 & 0.906 & 185.33 & 0.760 & 3.921& 9.221 & 0.046 & 0.014 & 10.16 & 0.367 & 4113.3 & 2.793 & 12.02 & 29.51 \\
    LoRA(CogX) & 6.151 & 0.928 & 126.25 &  0.654 &  3.712 & 6.405 & 0.043 & 0.008 & 9.983 & 0.371 & 2415.8 & 4.881 & 12.58 &  42.51\\
    TAV & 7.710 & 0.891 & 178.10 & 0.633 & 4.420 & 6.693 & \textcolor{blue}{\textbf{0.027}} & 0.007 & 10.17 & 0.360 & 2812.0 & \textcolor{red}{\textbf{2.455}} & 9.752 & 21.09 \\
    SimDA & 7.682 & 0.883 & 180.45 & 0.667 & 4.012 & \textcolor{blue}{\textbf{6.721}} & 0.031 & \textcolor{blue}{\textbf{0.006}} & 10.15 & 0.357 & 2965.1 & 2.832 & 9.212 & 25.61 \\
         \rowcolor{tabhighlight}
    Ours(SVD) & \textcolor{red}{\textbf{5.852}} & \textcolor{red}{\textbf{0.948}} & \textcolor{red}{\textbf{62.456}} & \textcolor{red}{\textbf{0.303}} & \textcolor{red}{\textbf{2.631}} & \textcolor{red}{\textbf{4.681}} & \textcolor{red}{\textbf{0.013}} & \textcolor{red}{\textbf{0.003}} & \textcolor{red}{\textbf{9.890}} & \textcolor{red}{\textbf{0.411}} & \textcolor{red}{\textbf{1742.5}} & \textcolor{blue}{\textbf{2.486}} & \textcolor{blue}{\textbf{8.160}} & \textcolor{blue}{\textbf{20.25}} \\
            \rowcolor{tabhighlight}
    Ours(CogX) & \textcolor{blue}{\textbf{5.946}} & \textcolor{blue}{\textbf{0.944}} & \textcolor{blue}{\textbf{124.81}} &  \textcolor{blue}{\textbf{0.608}} & \textcolor{blue}{\textbf{2.931}} & 8.508 & 0.040 & \textcolor{blue}{\textbf{0.006}} & \textcolor{blue}{\textbf{9.954}}  & \textcolor{blue}{\textbf{0.381}} & \textcolor{blue}{\textbf{1992.8}} & 2.553 & \textcolor{red}{\textbf{7.058}} & \textcolor{red}{\textbf{13.06}} \\

    \midrule
    \multicolumn{9}{c|}{\textbf{\textit{DamBreak fluid simulation}}} &
    \multicolumn{6}{c}{\textbf{\textit{Typhoon 202102}}}   \\ \midrule
    SVD & 8.189 & 0.743 & 1388.6 & 7.686 & 9.483 & 51.69 & 0.539 & 0.205 & 10.08 & 0.250 & 3290.9 & 8.953 & 17.99 & \textcolor{red}{\textbf{20.35}} \\ 
    LoRA(SVD) & 5.812 & 0.786 & 
    392.37 & 1.233 & 6.471 & 15.97 & 0.257 & 0.021 & 9.981 & 0.319 & 6264.3 & 17.94 & 16.48 & 49.02 \\
    LoRA(CogX) & 5.901 &  0.791 & 311.89 &   1.649 & 4.357 & 17.65 & 0.122 & 0.036 & 9.965 & 0.357 & 2179.5 & 4.353 & 12.36 & 42.72  \\
    TAV & 7.792 & 0.686 & 1543.6 & 2.572 & 10.08 & 34.33 & 0.205 & 0.140 & 10.07 & 0.298 & 3440.6 & \textcolor{blue}{\textbf{3.429}} & 14.46 & 25.95 \\
    SimDA & 7.679 & 0.691 & 1243.5 & 2.667 & 9.892 & 36.55 & 0.281 & 0.161 & 10.06 & 0.308 & 2898.5 & 3.509 & 14.01 & 26.07 \\
    \rowcolor{tabhighlight}
    Ours(SVD) & \textcolor{blue}{\textbf{4.921}} & \textcolor{blue}{\textbf{0.862}} & \textcolor{blue}{\textbf{158.36}} & \textcolor{blue}{\textbf{0.701}} & \textcolor{blue}{\textbf{3.271}} & \textcolor{blue}{\textbf{12.32}} & \textcolor{blue}{\textbf{0.032}} & \textcolor{blue}{\textbf{0.012}} & \textcolor{red}{\textbf{9.931}} & \textcolor{blue}{\textbf{0.371}} & \textcolor{red}{\textbf{1111.4}} & \textcolor{red}{\textbf{2.514}} & \textcolor{red}{\textbf{9.262}} & \textcolor{blue}{\textbf{23.87}} \\
    Ours(CogX) & \textcolor{red}{\textbf{4.492}} & \textcolor{red}{\textbf{0.884}} & \textcolor{red}{\textbf{88.493}} &  \textcolor{red}{\textbf{0.231}} & \textcolor{red}{\textbf{1.219}} & \textcolor{red}{\textbf{1.198}} &  \textcolor{red}{\textbf{0.009}} &  \textcolor{red}{\textbf{0.003}} & \textcolor{blue}{\textbf{9.903}} & \textcolor{red}{\textbf{0.374}} & \textcolor{blue}{\textbf{1149.4}} & 4.391  & \textcolor{blue}{\textbf{10.25}} &  33.72\\
    
    \midrule
    \multicolumn{9}{c|}{\textbf{\textit{DepthCharge fluid simulation}}} &
    \multicolumn{6}{c}{\textbf{\textit{Typhoon 202204}}} \\ \midrule
    SVD & 9.745 & 0.690 & 899.84 & 6.716 & 11.04 & 48.52 & 
    0.488 & 0.236 & 9.760 & 0.358 & 2624.2 & 7.883 & 15.78 & 39.09 \\ 
    LoRA(SVD) & 7.916 & 0.733 & 2213.9 & 7.405 & 13.19 & 75.91 & 1.300 & 0.482 & 9.677 & 0.338 & 1860.4 & 4.660 & 14.52 & \textcolor{red}{\textbf{22.66}} \\
    LoRA(CogX) &  4.775 & 0.812 &   212.26 & 1.366 & 5.492 & \textcolor{blue}{\textbf{12.04}} &  0.113 &   0.050 & 9.828 & 0.357 & 2937.1 & 4.785 & 14.86 & 39.46 \\
    
    TAV & 7.679 & 0.734 & 1307.7 & 
    5.777 & 13.04 & 61.66 & 0.587 & 0.256 & 9.649 & 0.359 & 2357.3 & 7.702 & 14.67 & 32.52 \\
    SimDA & 7.682 & 0.745 & 1298.2 & 5.867 & 14.01 & 67.21 & 0.576 & 0.301 &  9.686 &  0.347 &  2657.1 &  7.919 & 15.12&  31.89 \\
     \rowcolor{tabhighlight}
    Ours(SVD) & \textcolor{blue}{\textbf{4.672}} & \textcolor{blue}{\textbf{0.832}} & \textcolor{blue}{\textbf{418.18}} & \textcolor{blue}{\textbf{0.875}} & \textcolor{blue}{\textbf{8.026}} & 12.83 & \textcolor{blue}{\textbf{0.050}} & \textcolor{blue}{\textbf{0.025}} & \textcolor{blue}{\textbf{9.610}}& \textcolor{red}{\textbf{0.406}} & \textcolor{red}{\textbf{1462.8}} & \textcolor{red}{\textbf{2.830}} & \textcolor{blue}{\textbf{10.71}} & \textcolor{blue}{\textbf{29.48}} \\
         \rowcolor{tabhighlight}
    Ours(CogX) & \textcolor{red}{\textbf{3.990}} & \textcolor{red}{\textbf{0.872}} &  \textcolor{red}{\textbf{94.474}} & \textcolor{red}{\textbf{0.238}} & \textcolor{red}{\textbf{1.179}} & \textcolor{red}{\textbf{1.454}} &  \textcolor{red}{\textbf{0.011}} &  \textcolor{red}{\textbf{0.003}} & \textcolor{red}{\textbf{9.614}} & \textcolor{blue}{\textbf{0.396}} & \textcolor{blue}{\textbf{1608.8}} & \textcolor{blue}{\textbf{3.702}}  & \textcolor{red}{\textbf{11.39}} & 31.29 \\
    \bottomrule
    \end{tabular}}
    \caption{Quantitative evaluation on fluid simulation dataset (Left) and true typhoon dataset (Right). SVD: Stable Video Diffusion~\cite{blattmann2023stable}. CogX: CogVideoX~\cite{yang2024cogvideox} LoRA(SVD): SVD+LoRA~\cite{hu2021lora}. LoRA(CogX): SVD+CogX~\cite{yang2024cogvideox}. TAV: Tune-A-video~\cite{wu2023tune}. SimDA: Simple Diffusion Adapter~\cite{xing2023simda}.}
    \label{tab1}
\end{table*}

\textbf{Datasets and Evaluation Metrics.} We generate four simulated fluid scenarios using a computational fluid dynamics (CFD) toolkit: Rayleigh-Bénard Convection (RBC), Cylinder flow, DamBreak, and DepthCharge. To validate performance on real-world data, we further randomly select four typhoon events from true Typhoon dataset~\cite{digitaltyphoon}, identified by observation timestamps: 202001, 202009, 202102, and 202204. Formally, in each phenomenon test, we construct 10 videos, using 9 for fine-tuning and 1 for testing.
We employ eight metrics to evaluate both numerical accuracy and scientific fidelity: RMSE, SSIM~\cite{SSIM}, Stream Function Error (SFE)~\cite{StreanmFunc,Farneback}, Smoothness Error (SE)~\cite{Smoothness}, Gradient Smoothness (GS), Continuity Score (CS)~\cite{StreanmFunc}, Q-Criterion Error (QCE)~\cite{Q}, and Vorticity Error (VE)~\cite{Vorticity}. RMSE and SSIM measure per-pixel error and structural similarity, respectively. The remaining metrics are tailored for physical dynamics: SFE tests streamline accuracy, SE quantifies propagation smoothness, GS and CS (without ground truth) assess spatial-temporal consistency, and QCE and VE measure vortex preservation and vorticity accuracy. Notably, QCE and VE apply only to incompressible 2D flow fields from simulation data. Higher SSIM and lower scores on the other metrics indicate better performance. 

\textbf{Implementation Details.} Our model leverages the pre-trained SVD, the pre-trained CogVideoX, and the CLIP ViT-B/16 image encoder for visual conditioning. Fine-tuning is performed using SGD with a learning rate of $2 \times 10^{-4}$ for 15 epochs on NVIDIA A100 GPUs. A DDIM sampler~\cite{song2020denoising} with 50 steps is used during inference. The LoRA adaptation rank is set to 4, and frames are generated at a resolution of $512 \times 512$. The MAE and OFP are trained separately with a learning rate of $1 \times 10^{-3}$, batch size 60, and for 50 epochs each.

\subsection{Experimental Results}
\textbf{Qualitative Evaluation.} To visually assess the performance, we compare it against recent advanced approaches. Qualitative results are shown in Figure~\ref{fig3} (simulation fluid dynamic) and Figure~\ref{fig4} (real-world typhoon). In line with prior work~\cite{wu2023tune,xing2023simda}, we use the initial frame as a conditioning signal by concatenating it with the input noise. On simulated fluid data (Figure~\ref{fig3}), baseline models such as SVD without latent knowledge often generate unrealistic artifacts and hallucinations, even after fine-tuning. For example, in the DamBreak scenario, generated frames depict incorrect fluid behavior, including unnatural stillness or upward motion. By contrast, our approach, guided by latent scientific knowledge, produces phenomena that align better with underlying scientific principles.
A similar trend is observed in the real-world typhoon dataset (Figure~\ref{fig4}). SVD frequently produces hallucinated content that violates scientific plausibility. While methods like LoRA and TAV partially improve stability, they still introduce implausible distortions—such as transforming typhoon clouds into textures resembling snow-covered mountains or leather. In contrast, our model demonstrates a more accurate interpretation of input frame and generates temporally coherent sequences that reflect physically meaningful progression, highlighting the benefit of embedding latent scientific priors.

\textbf{Quantitative Evaluation.} We further evaluate our method using both numerical and physics-aware scientific metrics, as summarized in Table~\ref{tab1}. Overall, our method outperforms across most scenarios. On simulation datasets, our method achieves the best and the second-best performance in most metrics, while our method based on CogVideoX does not achieve top results in two scientific metrics, the results on other metrics still demonstrate its superiority. More importantly, the substantial improvements in scientific metrics achieved by our method, often by an order of magnitude, confirm its ability to generate sequences that better conform to fluid dynamics.
On the more complex typhoon dataset, our model continues to deliver state-of-the-art results across nearly all metrics. The only exception is the CS metric, which measures internal consistency without access to ground truth. Here, our performance is slightly weaker, likely due to the residual bias from pre-trained diffusion models toward nature-image coherence. When fine-tuned with limited data and physically grounded priors, this bias may not be fully corrected—pointing to a promising direction for future research.

\begin{table}[t!]
    % \scriptsize
    \footnotesize
    \centering
    \renewcommand{\arraystretch}{0.8}
    \renewcommand{\tabcolsep}{1.6mm}
    \scalebox{0.75}{
    \begin{tabular}{lcccccccc}
    \toprule
    & RMSE$\downarrow$ & SSIM$\uparrow$ & SFE$\downarrow$ & SE$\downarrow$ & GS$\downarrow$ & CS$\downarrow$ & QCE$\downarrow$ & VE$\downarrow$  \\ 
    \midrule
    % \multicolumn{9}{c|}{\textbf{\textit{RBC fluid simulation}}} \\ \midrule
    w/o SPK & 6.021 & 0.938 & 73.368 & 0.373 & 2.821 & 4.966 & 0.028 & 0.004   \\ 
    w/o DPK & 5.983 & 0.937 & 68.446 & 0.401 & 2.791 & 4.786 & 0.021 & 0.003   \\ 
    w/o SDM & 6.423 & 0.928 & 76.563 & 0.411 & 3.012 & 5.128 & 0.043 & 0.004   \\ 
    w/o FDM & 5.923 & 0.941 & 67.463 & 0.343 & 2.721 & 4.923 & 0.016 & 0.003   \\ 
    w/o QN & 5.913 & 0.940 & 73.161 & 0.398 & 2.823 & 4.823 & 0.018 & 0.004   \\
        \rowcolor{tabhighlight}
    Ours & \textbf{5.852} & \textbf{0.948}& \textbf{62.456} & \textbf{0.303} & \textbf{2.631} & \textbf{4.681} & \textbf{0.013} & \textbf{0.003}
    \\
    \bottomrule
    \end{tabular}}
    \caption{Ablation study on different components. w/o: without, SPK: static physical phenomenon knowledge, DPK: dynamic physical phenomenon knowledge, SDM: spatial domain modeling, FDM: frequency domain modeling, QN: quaternion network. w/o QN denotes replacing quaternion network with linear network. The best result is bolded.}
    \label{tab2}
\end{table}

\begin{table}[t!]
    % \scriptsize
    \footnotesize
    \centering
    \renewcommand{\arraystretch}{0.8}
    \renewcommand{\tabcolsep}{1.6mm}
    \scalebox{0.75}{
    \begin{tabular}{lcccccccc}
    \toprule
    Rank & RMSE$\downarrow$ & SSIM$\uparrow$ & SFE$\downarrow$ & SE$\downarrow$ & GS$\downarrow$ & CS$\downarrow$ & QCE$\downarrow$ & VE$\downarrow$  \\ 
    \midrule
    % \multicolumn{9}{c|}{\textbf{\textit{RBC fluid simulation}}} \\ \midrule
   2  & 6.132 & 0.928 & 78.213 & 0.376 & 2.798 & 4.945 & 0.034 & 0.004   \\ 

        \rowcolor{tabhighlight}
    4  & \textbf{5.852} & \textbf{0.948}& \textbf{62.456} & \textbf{0.303} & \textbf{2.631} & \textbf{4.681} & \textbf{0.013} & \textbf{0.003}
    \\
   8  & 6.018 & 0.936 & 68.356 & 0.351 & 2.756 & 5.084 & 0.027 & 0.003   \\
    \bottomrule
    \end{tabular}}
    \caption{Ablation study of LoRA rank.}
    \label{tab3}
\end{table}

\begin{table}[t!]
    % \scriptsize
    \footnotesize
    \centering
    \renewcommand{\arraystretch}{0.8}
    \renewcommand{\tabcolsep}{1.4mm}
    \scalebox{0.75}{
    \begin{tabular}{lcccccccc}
    \toprule
    Projection Methods & RMSE$\downarrow$ & SSIM$\uparrow$ & SFE$\downarrow$ & SE$\downarrow$ & GS$\downarrow$ & CS$\downarrow$ & QCE$\downarrow$ & VE$\downarrow$  \\ 
    \midrule
    % \multicolumn{9}{c|}{\textbf{\textit{RBC fluid simulation}}} \\ \midrule
   Linear Network  &5.913 &0.940 &73.161 &0.398 &2.823 &4.823 &0.018 &0.004   \\ 

   Cross Attention &5.921 &0.941 &70.012 &0.358 &2.705 &4.754 &0.016 &0.004
    \\
  \rowcolor{tabhighlight} Quaternion Network  &\textbf{5.852} &\textbf{0.948} &\textbf{62.456} &\textbf{0.303} &\textbf{2.631} &\textbf{4.681} &\textbf{0.013} &\textbf{0.003}   \\
    \bottomrule
    \end{tabular}}
    \caption{Ablation study of projection method.}
    \label{tab4}
\end{table}

\subsection{Ablation Study}
To further assess the effectiveness of individual components in our framework, we conduct a series of ablation studies on the Cylinder fluid simulation dataset. The results are summarized in Table~\ref{tab2}. We systematically remove key modules—such as static or dynamic scientific phenomenon knowledge and quaternion-based modeling—to evaluate their respective contributions. The ablation results demonstrate that each component plays a critical role. In particular, incorporating both static and dynamic latent scientific priors, as well as jointly modeling spatial and frequency representations via quaternion networks, significantly enhances the model’s ability to generate scientifically plausible sequences consistent with fundamental laws of motion.
We also explore the effect of varying the LoRA rank, with results reported in Table~\ref{tab3}. The performance is sensitive to this hyperparameter, indicating that careful selection is essential. For all main experiments, we use the rank that achieves the most stable and consistent results.
Finally, we evaluate different projection strategies for generating pseudo-language prompt embeddings. As shown in Table~\ref{tab4}, quaternion network-based projection yields better performance than alternative designs. This highlights the advantage of quaternion representations in capturing cross-domain semantics more effectively.

\section{Conclusion}
We present a novel framework that teaches video diffusion models with latent scientific phenomenon knowledge to enable more plausible generation from a single initial frame. Unlike existing approaches that rely heavily on natural vision priors and language prompts, our method leverages self-supervised learning to extract static knowledge and dynamic knowledge via optical flow prediction. We further propose a quaternion-based projection mechanism to convert latent knowledge into pseudo-language prompt embeddings, capturing both spatial and frequency-domain semantics. These embeddings are seamlessly integrated into video diffusion models through parameter-efficient fine-tuning, allowing the model to internalize latent knowledge and generalize to scientific phenomena. Extensive experiments on both simulated and real-world datasets demonstrate that our method potentially improves the scientific plausibility of generated videos, marking a step forward in bridging the gap between generative video models and scientific phenomena. 

\bibliography{aaai2026}

@article{ho2022video,
  title={Video diffusion models},
  author={Ho, Jonathan and Salimans, Tim and Gritsenko, Alexey and Chan, William and Norouzi, Mohammad and Fleet, David J},
  journal={Advances in Neural Information Processing Systems},
  year={2022}
}

@article{singer2022make,
  title={Make-a-video: Text-to-video generation without text-video data},
  author={Singer, Uriel and Polyak, Adam and Hayes, Thomas and Yin, Xi and An, Jie and Zhang, Songyang and Hu, Qiyuan and Yang, Harry and Ashual, Oron and Gafni, Oran and others},
  journal={arXiv:2209.14792},
  year={2022}
}

@inproceedings{blattmann2023align,
  title={Align your latents: High-resolution video synthesis with latent diffusion models},
  author={Blattmann, Andreas and Rombach, Robin and Ling, Huan and Dockhorn, Tim and Kim, Seung Wook and Fidler, Sanja and Kreis, Karsten},
  booktitle={Proceedings of the IEEE/CVF Conference on Computer Vision and Pattern Recognition},
  year={2023}
}

@article{ho2022imagen,
  title={Imagen video: High definition video generation with diffusion models},
  author={Ho, Jonathan and Chan, William and Saharia, Chitwan and Whang, Jay and Gao, Ruiqi and Gritsenko, Alexey and Kingma, Diederik P and Poole, Ben and Norouzi, Mohammad and Fleet, David J and others},
  journal={arXiv:2210.02303},
  year={2022}
}

@inproceedings{wu2023tune,
  title={Tune-a-video: One-shot tuning of image diffusion models for text-to-video generation},
  author={Wu, Jay Zhangjie and Ge, Yixiao and Wang, Xintao and Lei, Stan Weixian and Gu, Yuchao and Shi, Yufei and Hsu, Wynne and Shan, Ying and Qie, Xiaohu and Shou, Mike Zheng},
  booktitle={Proceedings of the IEEE/CVF International Conference on Computer Vision},
  year={2023}
}

@article{blattmann2023stable,
  title={Stable video diffusion: Scaling latent video diffusion models to large datasets},
  author={Blattmann, Andreas and Dockhorn, Tim and Kulal, Sumith and Mendelevitch, Daniel and Kilian, Maciej and Lorenz, Dominik and Levi, Yam and English, Zion and Voleti, Vikram and Letts, Adam and others},
  journal={arXiv:2311.15127},
  year={2023}
}

@inproceedings{he2022masked,
  title={Masked autoencoders are scalable vision learners},
  author={He, Kaiming and Chen, Xinlei and Xie, Saining and Li, Yanghao and Doll{\'a}r, Piotr and Girshick, Ross},
  booktitle={Proceedings of the IEEE/CVF Conference on Computer Vision and Pattern Recognition},
  year={2022}
}

@inproceedings{radford2021learning,
  title={Learning transferable visual models from natural language supervision},
  author={Radford, Alec and Kim, Jong Wook and Hallacy, Chris and Ramesh, Aditya and Goh, Gabriel and Agarwal, Sandhini and Sastry, Girish and Askell, Amanda and Mishkin, Pamela and Clark, Jack and others},
  booktitle={International Conference on Machine Learning},
  year={2021},
}

@article{parcollet2018quaternion,
  title={Quaternion recurrent neural networks},
  author={Parcollet, Titouan and Ravanelli, Mirco and Morchid, Mohamed and Linar{\`e}s, Georges and Trabelsi, Chiheb and De Mori, Renato and Bengio, Yoshua},
  journal={arXiv:1806.04418},
  year={2018}
}

@article{song2020denoising,
  title={Denoising diffusion implicit models},
  author={Song, Jiaming and Meng, Chenlin and Ermon, Stefano},
  journal={arXiv:2010.02502},
  year={2020}
}

@article{ho2020denoising,
  title={Denoising diffusion probabilistic models},
  author={Ho, Jonathan and Jain, Ajay and Abbeel, Pieter},
  journal={Advances in Neural Information Processing Systems},
  year={2020}
}

@article{croitoru2023diffusion,
  title={Diffusion models in vision: A survey},
  author={Croitoru, Florinel-Alin and Hondru, Vlad and Ionescu, Radu Tudor and Shah, Mubarak},
  journal={IEEE Transactions on Pattern Analysis and Machine Intelligence},
  year={2023},
}

@article{ho2022classifier,
  title={Classifier-free diffusion guidance},
  author={Ho, Jonathan and Salimans, Tim},
  journal={arXiv:2207.12598},
  year={2022}
}

@article{karras2022elucidating,
  title={Elucidating the design space of diffusion-based generative models},
  author={Karras, Tero and Aittala, Miika and Aila, Timo and Laine, Samuli},
  journal={Advances in Neural Information Processing Systems},
  year={2022}
}

@article{salimans2022progressive,
  title={Progressive distillation for fast sampling of diffusion models},
  author={Salimans, Tim and Ho, Jonathan},
  journal={arXiv:2202.00512},
  year={2022}
}

@article{kawar2022denoising,
  title={Denoising diffusion restoration models},
  author={Kawar, Bahjat and Elad, Michael and Ermon, Stefano and Song, Jiaming},
  journal={Advances in Neural Information Processing Systems},
  year={2022}
}

@article{li2022srdiff,
  title={Srdiff: Single image super-resolution with diffusion probabilistic models},
  author={Li, Haoying and Yang, Yifan and Chang, Meng and Chen, Shiqi and Feng, Huajun and Xu, Zhihai and Li, Qi and Chen, Yueting},
  journal={Neurocomputing},
  year={2022},
}

@inproceedings{lugmayr2022repaint,
  title={Repaint: Inpainting using denoising diffusion probabilistic models},
  author={Lugmayr, Andreas and Danelljan, Martin and Romero, Andres and Yu, Fisher and Timofte, Radu and Van Gool, Luc},
  booktitle={Proceedings of the IEEE/CVF Conference on Computer Vision and Pattern Recognition},
  year={2022}
}

@article{nichol2021glide,
  title={Glide: Towards photorealistic image generation and editing with text-guided diffusion models},
  author={Nichol, Alex and Dhariwal, Prafulla and Ramesh, Aditya and Shyam, Pranav and Mishkin, Pamela and McGrew, Bob and Sutskever, Ilya and Chen, Mark},
  journal={arXiv:2112.10741},
  year={2021}
}

@article{ramesh2022hierarchical,
  title={Hierarchical text-conditional image generation with clip latents},
  author={Ramesh, Aditya and Dhariwal, Prafulla and Nichol, Alex and Chu, Casey and Chen, Mark},
  journal={arXiv:2204.06125},
  year={2022}
}

@inproceedings{rombach2022high,
  title={High-resolution image synthesis with latent diffusion models},
  author={Rombach, Robin and Blattmann, Andreas and Lorenz, Dominik and Esser, Patrick and Ommer, Bj{\"o}rn},
  booktitle={Proceedings of the IEEE/CVF Conference on Computer Vision and Pattern Recognition},
  year={2022}
}

@article{saharia2022photorealistic,
  title={Photorealistic text-to-image diffusion models with deep language understanding},
  author={Saharia, Chitwan and Chan, William and Saxena, Saurabh and Li, Lala and Whang, Jay and Denton, Emily L and Ghasemipour, Kamyar and Gontijo Lopes, Raphael and Karagol Ayan, Burcu and Salimans, Tim and others},
  journal={Advances in Neural Information Processing Systems},
  year={2022}
}

@inproceedings{khachatryan2023text2video,
  title={Text2video-zero: Text-to-image diffusion models are zero-shot video generators},
  author={Khachatryan, Levon and Movsisyan, Andranik and Tadevosyan, Vahram and Henschel, Roberto and Wang, Zhangyang and Navasardyan, Shant and Shi, Humphrey},
  booktitle={Proceedings of the IEEE/CVF International Conference on Computer Vision},
  year={2023}
}

@inproceedings{li2024vidtome,
  title={Vidtome: Video token merging for zero-shot video editing},
  author={Li, Xirui and Ma, Chao and Yang, Xiaokang and Yang, Ming-Hsuan},
  booktitle={Proceedings of the IEEE/CVF Conference on Computer Vision and Pattern Recognition},
  year={2024}
}

@inproceedings{xing2024simda,
  title={Simda: Simple diffusion adapter for efficient video generation},
  author={Xing, Zhen and Dai, Qi and Hu, Han and Wu, Zuxuan and Jiang, Yu-Gang},
  booktitle={Proceedings of the IEEE/CVF Conference on Computer Vision and Pattern Recognition},
  year={2024}
}

@inproceedings{jiang2024videobooth,
  title={Videobooth: Diffusion-based video generation with image prompts},
  author={Jiang, Yuming and Wu, Tianxing and Yang, Shuai and Si, Chenyang and Lin, Dahua and Qiao, Yu and Loy, Chen Change and Liu, Ziwei},
  booktitle={Proceedings of the IEEE/CVF Conference on Computer Vision and Pattern Recognition},
  year={2024}
}

@inproceedings{zhang2023adding,
  title={Adding conditional control to text-to-image diffusion models},
  author={Zhang, Lvmin and Rao, Anyi and Agrawala, Maneesh},
  booktitle={Proceedings of the IEEE/CVF International Conference on Computer Vision},
  year={2023}
}

@article{zhang2023controlvideo,
  title={Controlvideo: Training-free controllable text-to-video generation},
  author={Zhang, Yabo and Wei, Yuxiang and Jiang, Dongsheng and Zhang, Xiaopeng and Zuo, Wangmeng and Tian, Qi},
  journal={ arXiv:2305.13077},
  year={2023}
}

@inproceedings{saharia2022palette,
  title={Palette: Image-to-image diffusion models},
  author={Saharia, Chitwan and Chan, William and Chang, Huiwen and Lee, Chris and Ho, Jonathan and Salimans, Tim and Fleet, David and Norouzi, Mohammad},
  booktitle={ACM SIGGRAPH 2022 conference proceedings},
  year={2022}
}

@article{kazerouni2022diffusion,
  title={Diffusion models for medical image analysis: A comprehensive survey},
  author={Kazerouni, Amirhossein and Aghdam, Ehsan Khodapanah and Heidari, Moein and Azad, Reza and Fayyaz, Mohsen and Hacihaliloglu, Ilker and Merhof, Dorit},
  journal={arXiv:2211.07804},
  year={2022}
}

@inproceedings{sohl2015deep,
  title={Deep unsupervised learning using nonequilibrium thermodynamics},
  author={Sohl-Dickstein, Jascha and Weiss, Eric and Maheswaranathan, Niru and Ganguli, Surya},
  booktitle={International Conference on Machine Learning},
  year={2015},
}

@article{song2020score,
  title={Score-based generative modeling through stochastic differential equations},
  author={Song, Yang and Sohl-Dickstein, Jascha and Kingma, Diederik P and Kumar, Abhishek and Ermon, Stefano and Poole, Ben},
  journal={arXiv:2011.13456},
  year={2020}
}

@article{hu2021lora,
  title={Lora: Low-rank adaptation of large language models},
  author={Hu, Edward J and Shen, Yelong and Wallis, Phillip and Allen-Zhu, Zeyuan and Li, Yuanzhi and Wang, Shean and Wang, Lu and Chen, Weizhu},
  journal={arXiv:2106.09685},
  year={2021}
}

@article{dosovitskiy2020image,
  title={An image is worth 16x16 words: Transformers for image recognition at scale},
  author={Dosovitskiy, Alexey},
  journal={arXiv:2010.11929},
  year={2020}
}

@article{SSIM,
  title={Image quality assessment: from error visibility to structural similarity},
  author={Zhou Wang and Alan Conrad Bovik and Hamid R. Sheikh and Eero P. Simoncelli},
  journal={IEEE Transactions on Image Processing},
  year={2004},
}

@inproceedings{Farneback,
  title={Two-Frame Motion Estimation Based on Polynomial Expansion},
  author={Gunnar Farneb{\"a}ck},
  booktitle={Scandinavian Conference on Image Analysis},
  year={2003},
}

@book{StreanmFunc,
  title={Fluid Mechanics},
  author={Kundu, Pijush K. and Cohen, Ira M. and Dowling, David R.},
  edition={6},
  year={2015},
  publisher={Elsevier}
}

@INPROCEEDINGS{Q,
  author = {{Hunt}, J.~C.~R. and {Wray}, A.~A. and {Moin}, P.},
  title = "{Eddies, streams, and convergence zones in turbulent flows}",
  booktitle = {Studying Turbulence Using Numerical Simulation Databases, 2},
  year = 1988,
}

@article{Smoothness,
  title={Imposing Consistency for Optical Flow Estimation},
  author={Jisoo Jeong and Jamie Menjay Lin and Fatih Murat Porikli and Nojun Kwak},
  journal={Proceedings of the IEEE/CVF Conference on Computer Vision and Pattern Recognition},
  year={2022},
}

@article{Vorticity,
  title={On the identification of a vortex},
  author={Jinhee Jeong and Fazle Hussain},
  journal={Journal of Fluid Mechanics},
  year={1995},
}

@InProceedings{digitaltyphoon,
      author = {Asanobu Kitamoto and Jared Hwang and Bastien Vuillod and Lucas Gautier and Yingtao Tian and Tarin Clanuwat},
      title = {Digital Typhoon: Long-term Satellite Image Dataset for the Spatio-Temporal Modeling of Tropical Cyclones},
      booktitle = {NeurIPS 2023 Datasets and Benchmarks},
      year = 2023,
}

@inproceedings{xing2023simda,
  title={SimDA: Simple Diffusion Adapter for Efficient Video Generation},
  author={Xing, Zhen and Dai, Qi and Hu, Han and Wu, Zuxuan and Jiang, Yu-Gang},
  booktitle={Proceedings of the IEEE/CVF Conference on Computer Vision and Pattern Recognition},
  year={2024}
}

@inproceedings{cao2024domain,
  title={Domain prompt learning with quaternion networks},
  author={Cao, Qinglong and Xu, Zhengqin and Chen, Yuntian and Ma, Chao and Yang, Xiaokang},
  booktitle={Proceedings of the IEEE/CVF Conference on Computer Vision and Pattern Recognition},
  pages={26637--26646},
  year={2024}
}

@inproceedings{farneback2003two,
  title={Two-frame motion estimation based on polynomial expansion},
  author={Farneb{\"a}ck, Gunnar},
  booktitle={Image Analysis: 13th Scandinavian Conference, SCIA 2003 Halmstad, Sweden, June 29--July 2, 2003 Proceedings 13},
  pages={363--370},
  year={2003},
  organization={Springer}
}

@inproceedings{shiprompt,
  title={Prompt Learning with Quaternion Networks},
  author={Shi, Boya and Xu, Zhengqin and Jia, Shuai and Ma, Chao},
  booktitle={The Twelfth International Conference on Learning Representations},
  year={2024}
}

@article{yang2023learning,
  title={Learning interactive real-world simulators},
  author={Yang, Mengjiao and Du, Yilun and Ghasemipour, Kamyar and Tompson, Jonathan and Schuurmans, Dale and Abbeel, Pieter},
  journal={arXiv preprint arXiv:2310.06114},
  volume={1},
  number={2},
  pages={6},
  year={2023}
}

@article{kang2024far,
  title={How far is video generation from world model: A physical law perspective},
  author={Kang, Bingyi and Yue, Yang and Lu, Rui and Lin, Zhijie and Zhao, Yang and Wang, Kaixin and Huang, Gao and Feng, Jiashi},
  journal={arXiv preprint arXiv:2411.02385},
  year={2024}
}

@article{zhang2022dino,
  title={Dino: Detr with improved denoising anchor boxes for end-to-end object detection},
  author={Zhang, Hao and Li, Feng and Liu, Shilong and Zhang, Lei and Su, Hang and Zhu, Jun and Ni, Lionel M and Shum, Heung-Yeung},
  journal={arXiv preprint arXiv:2203.03605},
  year={2022}
}

@article{zhang2024filtered,
  title={Filtered partial differential equations: a robust surrogate constraint in physics-informed deep learning framework},
  author={Zhang, Dashan and Chen, Yuntian and Chen, Shiyi},
  journal={Journal of Fluid Mechanics},
  volume={999},
  pages={A40},
  year={2024},
  publisher={Cambridge University Press}
}

@article{chen2025three,
  title={Three-dimensional spatiotemporal wind field reconstruction based on LiDAR and multi-scale PINN},
  author={Chen, Yuanqing and Wang, Ding and Feng, Dachuan and Tian, Geng and Gupta, Vikrant and Cao, Renjing and Wan, Minping and Chen, Shiyi},
  journal={Applied Energy},
  volume={377},
  pages={124577},
  year={2025},
  publisher={Elsevier}
}

@article{yang2024cogvideox,
  title={Cogvideox: Text-to-video diffusion models with an expert transformer},
  author={Yang, Zhuoyi and Teng, Jiayan and Zheng, Wendi and Ding, Ming and Huang, Shiyu and Xu, Jiazheng and Yang, Yuanming and Hong, Wenyi and Zhang, Xiaohan and Feng, Guanyu and others},
  journal={arXiv preprint arXiv:2408.06072},
  year={2024}
}

@article{he2024cameractrl,
  title={Cameractrl: Enabling camera control for text-to-video generation},
  author={He, Hao and Xu, Yinghao and Guo, Yuwei and Wetzstein, Gordon and Dai, Bo and Li, Hongsheng and Yang, Ceyuan},
  journal={arXiv preprint arXiv:2404.02101},
  year={2024}
}

@article{he2025liangke,
  title={Liangke Gui, Qi Zhao, Gordon Wetzstein, Lu Jiang, and Hongsheng Li. Cameractrl ii: Dynamic scene exploration via camera-controlled video diffusion models},
  author={He, Hao and Yang, Ceyuan and Lin, Shanchuan and Xu, Yinghao and Wei, Meng},
  journal={arXiv preprint arXiv:2503.10592},
  volume={19},
  year={2025}
}

@article{zhang2025decouple,
  title={Decouple before align: Visual disentanglement enhances prompt tuning},
  author={Zhang, Fei and Zhou, Tianfei and Yao, Jiangchao and Zhang, Ya and Tsang, Ivor W and Wang, Yanfeng},
  journal={IEEE Transactions on Pattern Analysis and Machine Intelligence},
  year={2025},
  publisher={IEEE}
}

@article{zhou2024image,
  title={Image segmentation in foundation model era: A survey},
  author={Zhou, Tianfei and Xia, Wang and Zhang, Fei and Chang, Boyu and Wang, Wenguan and Yuan, Ye and Konukoglu, Ender and Cremers, Daniel},
  journal={arXiv preprint arXiv:2408.12957},
  year={2024}
}

@article{cao2025generalized,
  title={Generalized domain prompt learning for accessible scientific vision-language models},
  author={Cao, Qinglong and Chen, Yuntian and Lu, Lu and Sun, Hao and Zeng, Zhengzhong and Yang, Xiaokang and Zhang, Dongxiao},
  journal={Nexus},
  volume={2},
  number={2},
  year={2025},
  publisher={Elsevier}
}

@inproceedings{gao2025auto,
  title={Auto-regressive moving diffusion models for time series forecasting},
  author={Gao, Jiaxin and Cao, Qinglong and Chen, Yuntian},
  booktitle={Proceedings of the AAAI Conference on Artificial Intelligence},
  volume={39},
  number={16},
  pages={16727--16735},
  year={2025}
}

% Check whether the conference requires a reproducibility checklist to be included in the paper.
% If so, you can uncomment the following line and ajust the path to include it.
% \input{../../ReproducibilityChecklist/LaTeX/ReproducibilityChecklist.tex}

\clearpage
\appendix
\twocolumn[
\begin{center}
    \textbf{\Large Supplementary Material for Latent Knowledge-Guided Video Diffusion for Scientific Phenomena Generation
from a Single Initial Frame}
\end{center}
]

\section{Criteria}
\label{sec:criteria}

\textbf{Root Mean Squared Error (RMSE).} RMSE provides a comprehensive measure of pixel-level deviations between the generated and real videos. A lower RMSE indicates that the generated video is closer to the real video in terms of pixel values. The formula is as follows:
\begin{equation}
\mathrm{RMSE} = \frac{1}{N} \sum_{t=1}^{N} \sqrt{\frac{1}{M} \sum_{i=1}^{M} \left( X_i^{\mathrm{real}} - X_i^{\mathrm{gen}} \right)^2},
\end{equation}
where $N$ is the number of frames, and $M$ is the total number of pixels per frame. $<\cdot>^{\mathrm{real}}$ and $<\cdot>^{\mathrm{gen}}$ are true values and generated values respectively, as well as $X_i$ are pixel values. \\

\textbf{Structural Similarity Index Measure (SSIM).} 
SSIM \cite{SSIM} assesses structural similarity between generated and real video frames, focusing on aspects like brightness, contrast, and texture. This metric helps ensure that the generated video maintains the structural integrity of the real video. Higher SSIM values indicate greater similarity. The formula is:
\begin{equation}
\mathrm{SSIM} = \frac{(2\mu_{x}\mu_{y} + C_1)(2\sigma_{xy} + C_2)}{(\mu_{x}^2 + \mu_{y}^2 + C_1)(\sigma_{x}^2 + \sigma_{y}^2 + C_2)},
\end{equation}
where $\mu_{x}$ and $\mu_{y}$ are the mean values, $\sigma_{x}$ and $\sigma_{y}$ are the variances of frames $x$ and $y$, $\sigma_{xy}$ is the covariance, and $C_1$ and $C_2$ are constants to stabilize the division. \\

Additionally, to evaluate the generated video's quality, we analyze the velocity fields derived from optical flow based on physical quantities such as velocity, temperature, and volume fraction in the original video. This allows us to apply metrics that assess the alignment of the generated video with the original's physical characteristics. These metrics, grounded in convective dynamics, are suitable for analyzing the flow consistency across different physical quantities, determining if they exhibit smooth and realistic motion over time.

We compute the optical flow fields using the Farneback method \cite{Farneback}, a dense optical flow estimation technique that approximates each pixel's local neighborhood as a polynomial function. This approach facilitates precise, pixel-wise motion estimates, making it ideal for capturing fine-grained movements in fluid-related fields.

In Farneback’s method, each pixel neighborhood is represented by a quadratic polynomial:
\begin{equation}
f(\mathbf{x}) = \mathbf{x}^\top \mathbf{A} \mathbf{x} + \mathbf{b}^\top \mathbf{x} + c,
\end{equation}
where $\mathbf{x}$ is the pixel location, $\mathbf{A}$ represents second-order terms, $\mathbf{b}$ denotes first-order terms, and $c$ is a constant. The displacement vector $\mathbf{d}$, representing pixel-wise motion, is derived from changes in these terms across frames:
\begin{equation}
\mathbf{d} = -\left(\mathbf{A} + \mathbf{A}^\top\right)^{-1} \left(\mathbf{b} - \mathbf{b}^\prime\right).
\end{equation}
Using a pyramidal approach, the Farneback algorithm captures motion at multiple scales, refining the motion field at each level for accurate displacement estimation. The resulting optical flow vectors $\mathbf{u} = (u, v)$ provide the basis for calculating metrics that evaluate the generated video’s fidelity to the physical properties of the original. \\

\textbf{Stream Function Error (SFE).}
The stream function, $\psi$, represents a scalar field from which the velocity components of a two-dimensional incompressible flow can be derived, where $\partial \psi / \partial y = u$ and $\partial \psi / \partial x = -v$ \cite{StreanmFunc}. For optical flow fields, $\psi$ is computed via numerical integration:
\begin{equation}
\psi(x, y) = \int u \, \mathrm{d}y - \int v \, \mathrm{d}x.
\end{equation}
The Stream Function Error (SFE) between generated and real data is calculated as:
\begin{equation}
\mathrm{SFE} = \frac{1}{N} \sum_{t=1}^{N} \sqrt{\frac{1}{M} \sum_{i=1}^{M} \left( \psi_i^{\mathrm{real}} - \psi_i^{\mathrm{gen}} \right)^2}.
\end{equation}
SFE assesses the dynamic consistency of the generated flow field with the real flow by comparing streamline characteristics. Lower SFE values indicate that the generated flow better replicates advective properties, providing insights into the physical accuracy and quality of the generated video. \\

\textbf{Smoothness Error (SE).}
Smooth changes in velocity generally reflect the asymptotic behavior of physical phenomena, while abrupt fluctuations may be unrealistic \cite{Smoothness}. Temporal smoothness in optical flow velocity can capture the steady propagation characteristics of the underlying physical quantities. Smoothness Error (SE) measures the timewise smoothness of both the generated and real flow fields, providing insight into physical continuity over time. SE is defined as:
\begin{equation}
\mathrm{SE} = \frac{1}{N-1} \sum_{t=1}^{N-1} \sqrt{\frac{1}{M} \sum_{i=1}^{M} \left( \Delta u_i^{\mathrm{gen}} - \Delta u_i^{\mathrm{real}} \right)^2},
\end{equation}
where $\Delta u_i = u_{i,t+1} - u_{i,t}$ represents the velocity change across consecutive time intervals. A lower SE value indicates greater temporal smoothness in the generated flow, reflecting the essential physical continuity of the quantity. \\

 \begin{figure*}[ht!]
	\begin{center}
		%\fbox{\rule{0pt}{2in} \rule{0.9\linewidth}{0pt}}
		\includegraphics[width=1.0\linewidth]{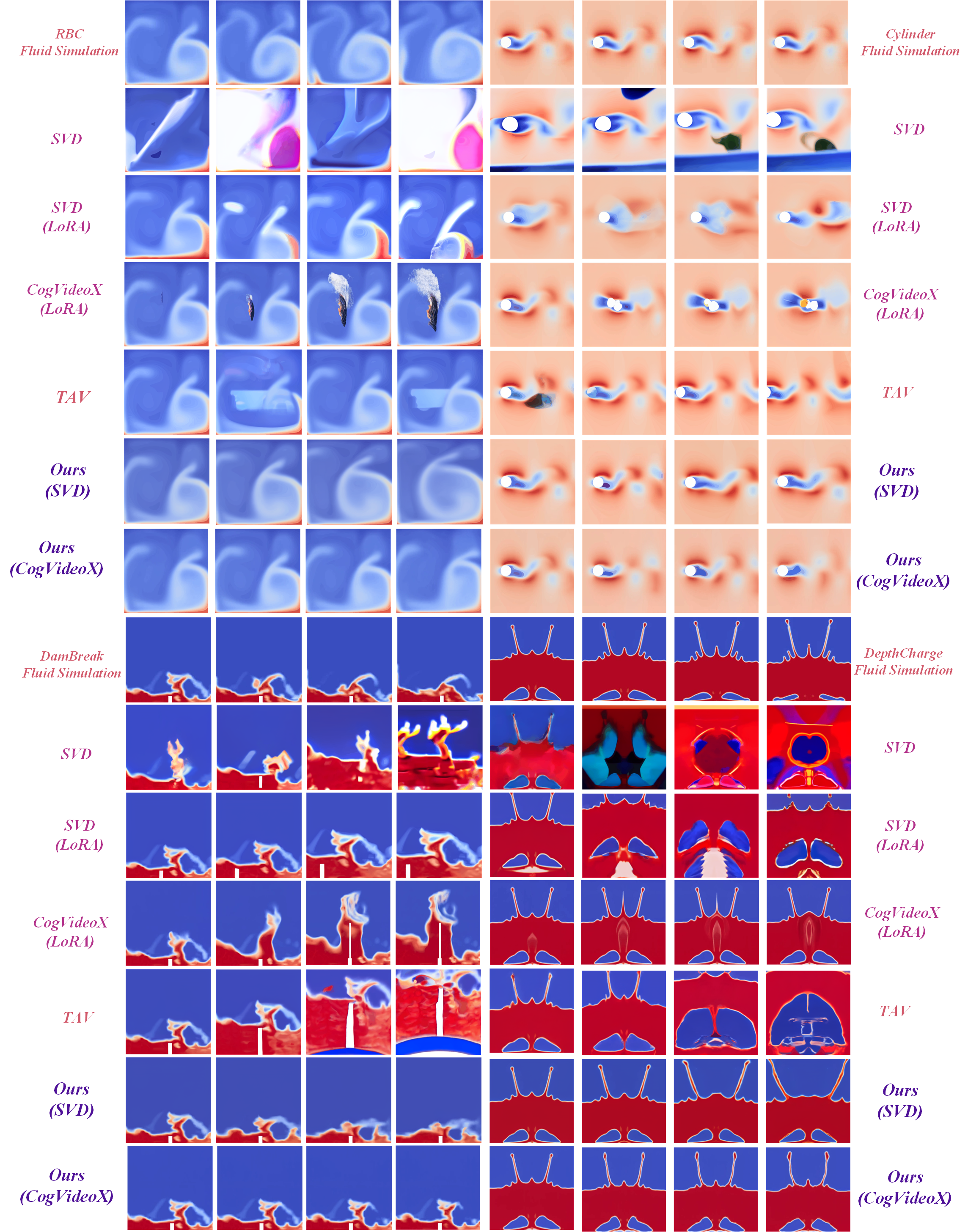}
	\end{center}
        % \vspace{-7mm}
	\caption{Qualitative comparisons in fluid simulation dataset. Though incorporating physical phenomenon knowledge, our method generates rational phenomena that exhibit better alignment with physical laws.  }
	\label{fig7}
    % \vspace{-5mm}
\end{figure*}

 \begin{figure*}[ht!]
	\begin{center}
		%\fbox{\rule{0pt}{2in} \rule{0.9\linewidth}{0pt}}
		\includegraphics[width=1.0\linewidth]{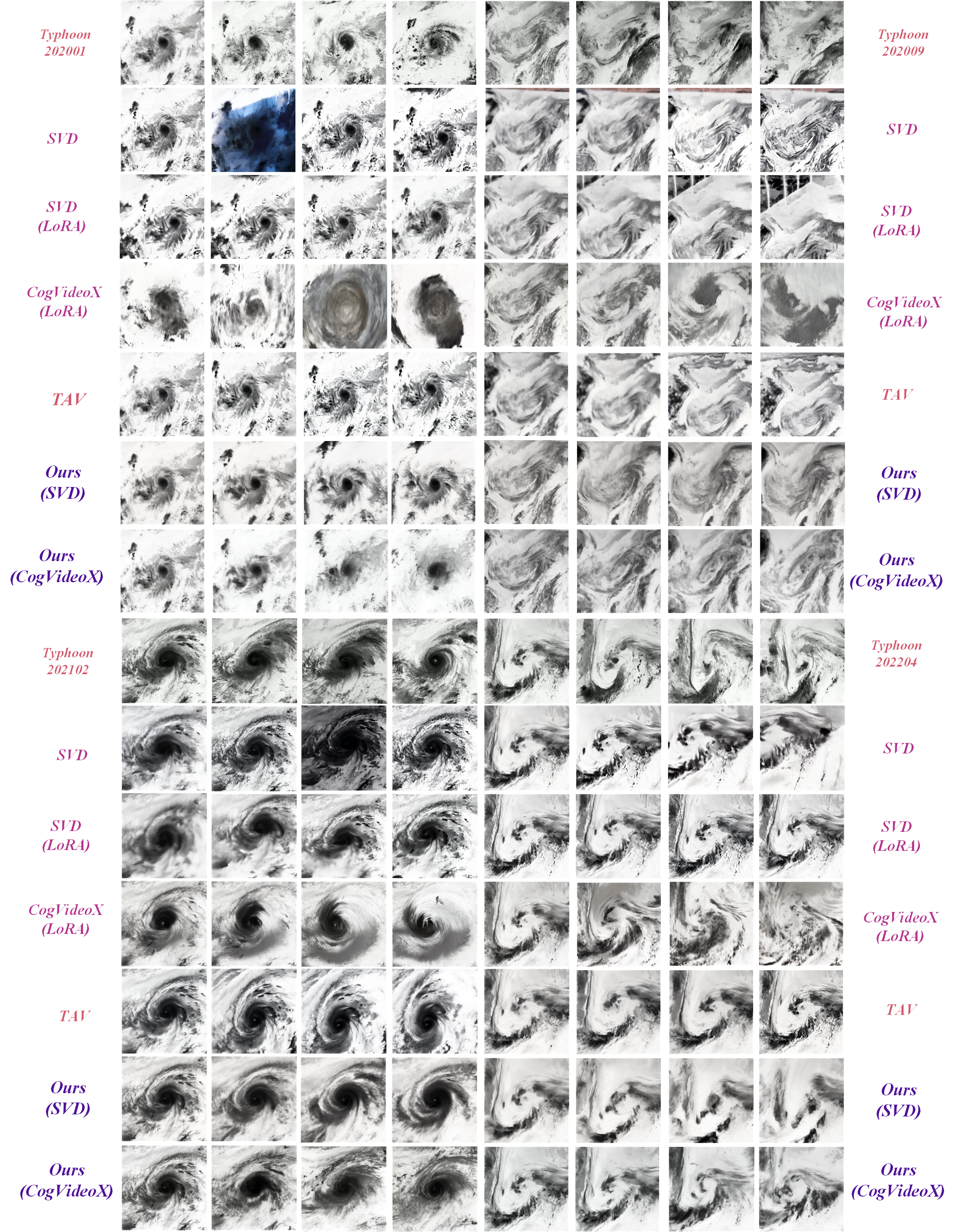}
	\end{center}
        % \vspace{-7mm}
	\caption{Qualitative comparisons in true typhoon dataset. Though incorporating physical phenomenon knowledge, our method generates rational phenomena that exhibit better alignment with physical laws.  }
	\label{fig8}
    % \vspace{-5mm}
\end{figure*}

\textbf{Gradient Smoothness (GS).}
Gradient Smoothness evaluates the temporal smoothness of the gradient field in the generated frames, capturing the physical continuity of spatial features across time steps. The formula is:
\begin{align}
\mathrm{GS} &= \frac{1}{N-1} \sum_{t=1}^{N-1} \nonumber \\
&\sqrt{\frac{\sum_{i=1}^{M} \left( \left( \frac{\partial X_i^{t+1}}{\partial x} - \frac{\partial X_i^t}{\partial x} \right)^2 + \left( \frac{\partial X_i^{t+1}}{\partial y} - \frac{\partial X_i^t}{\partial y} \right)^2 \right)}{2M}}.
\end{align}
This metric is computed solely from the generated frames and reflects the smoothness of changes in the gradient field over time. \\

\textbf{Continuity Score (CS).}
Continuity Score measures the spatial continuity of the generated optimal flow field by calculating the divergence of each frame in the generated sequence. The formula is:
\begin{equation}
\mathrm{CS} = \frac{1}{N} \sum_{t=1}^{N} \sqrt{\frac{1}{M} \sum_{i=1}^{M} \left( \nabla \cdot \mathbf{u}_i \right)^2},
\end{equation}
where $\nabla \cdot \mathbf{u} = \partial u/\partial x + \partial v/\partial y$ represents the divergence of the generated velocity field \( \mathbf{u} = (u, v) \) at each pixel \( i \) and time step \( t \). Lower CS values indicate better preservation of spatial continuity in the generated field, reflecting adherence to mass conservation principles. \\

\textbf{Q-Criterion Error (QCE).}
The Q-Criterion identifies vortices within a flow field by balancing rotational and strain rates \cite{Q}. It’s calculated as:
\begin{equation}
Q = \frac{1}{2} \left( \| \mathbf{\Omega} \|^2 - \| \mathbf{S} \|^2 \right),
\end{equation}
where $\mathbf{\Omega}= 1/2 \left( \nabla \mathbf{u} - (\nabla \mathbf{u})^T \right)$ is rotation tensor and $\mathbf{S} = 1/2 \left( \nabla \mathbf{u} + (\nabla \mathbf{u})^T \right)$ is strain tensor. In the two-dimensional plane, the calculation can be simplified as:
\begin{equation}
Q = \frac{1}{2} \left( -\left( \frac{\partial u}{\partial x} \right)^2 - \left( \frac{\partial v}{\partial y} \right)^2 - 2 \frac{\partial u}{\partial y} \frac{\partial v}{\partial x} \right),
\end{equation}
with error:
\begin{equation}
\mathrm{QCE} = \frac{1}{N} \sum_{t=1}^{N} \sqrt{\frac{1}{M} \sum_{i=1}^{M} \left( Q_i^{\mathrm{gen}} - Q_i^{\mathrm{real}} \right)^2}.
\end{equation}
Lower QCE suggests that the generated flow retains rotational structures similar to the real flow. \\

\textbf{Vorticity Error (VE).}
Vorticity represents rotational effects, such as eddies and vortices, which are vital for processes like mixing, energy transfer, and turbulence development \cite{Vorticity}. In two-dimensional flow, local rotation is quantified by vorticity, defined as: 
\begin{equation} 
\omega = \frac{\partial v}{\partial x} - \frac{\partial u}{\partial y}. \end{equation} 
The Vorticity Error (VE) measures the accuracy of the generated flow's rotational dynamics compared to the real flow and is calculated as: 
\begin{equation} \mathrm{VE} = \frac{1}{N} \sum_{t=1}^{N} \sqrt{\frac{1}{M} \sum_{i=1}^{M} \left( \omega_i^{\mathrm{gen}} - \omega_i^{\mathrm{real}} \right)^2}. 
\end{equation} 
Lower VE values indicate closer alignment in rotational characteristics, showing that the generated flow accurately replicates the real flow’s dynamics. 
% This metric is crucial for validating flow accuracy in scenarios where precise representation of turbulence and vorticity-driven phenomena, such as typhoons, is essential.

\section{CFD simulations}
Rayleigh-Bénard convection is a thermally-driven natural convection flow caused by temperature differences, which induces ascending and descending currents and is commonly used to study turbulence and heat transfer. Cylinder flow simulates the fluid behavior around an obstacle (e.g., a cylinder), producing wake and vortex structures to examine vortex dynamics in bluff-body flows. DamBreak flow models the propagation of free-surface waves and fluid flow following a sudden dam collapse, illustrating gravity-driven, nonlinear free-surface behavior. DepthCharge flow simulates the high-pressure shockwave and bubble expansion dynamics following an underwater explosion, focusing on the transient changes at the water-air interface.

The numerical simulations for these flows were conducted using the open-source software OpenFOAM, which employs the finite volume method for discretizing partial differential equations (PDEs). The Rayleigh-Bénard convection was modeled with a RANS turbulence approach, while the other flows were treated as laminar. A second-order central difference scheme was applied to the diffusion terms, and appropriate discretization schemes were selected for the convection terms according to the specific physical requirements of each flow. Temporal discretization used a second-order backward implicit method for Rayleigh-Bénard convection, while a first-order explicit Euler scheme was applied for the others. The simulations were conducted over 400 time steps, with results interpolated from nonuniform grids to a uniform spatial resolution of 512 × 512. The processed frames were then compiled into 16 videos.

\section{Qualitative Comparisons.} To provide a more comprehensive analysis of the performance of our proposed method, we present additional qualitative comparisons. The visualization results are shown in Figure~\ref{fig7} and Figure~\ref{fig8}, with Figure~\ref{fig7} illustrating the comparisons on the fluid simulation dataset and Figure~\ref{fig8} showing the results on the true typhoon dataset. These examples consistently highlight the superior performance of our method, showcasing its robustness and improved alignment with physical realism across various scenarios.

\section{Provided Videos.}To directly assess the performance of our proposed method, we also provide a comprehensive review video, titled ``Comparison Results.mp4", which summarizes the results. The video and additional detailed videos corresponding to specific experiments can be found in the supplementary.

\end{document}